\RequirePackage{fix-cm}

\documentclass[smallextended]{svjour3}    

\smartqed  
\usepackage[inline]{enumitem}
\usepackage{graphicx}
\usepackage{hyperref}
\usepackage[backend=biber,sorting=nyt,bibencoding=latin1,abbreviate=false,mincrossrefs=3,style=numeric,maxnames=30]{biblatex}
\usepackage{todonotes}

\begin{document}

\title{A Reference Software Architecture for Social Robots}

\author{Luigi Asprino \and Paolo Ciancarini \and Andrea Giovanni Nuzzolese \and Valentina Presutti \and Alessandro Russo}

\authorrunning{Asprino et al.}

\institute{L. Asprino \at 
FICLIT, University of Bologna, via Zamboni, 32 - 40126 - Bologna \\
STLab (ISTC-CNR), Via San Martino della Battaglia, 44, 00185 Rome, Italy \\
\email{luigi.asprino@unibo.it}\\
\and A. G. Nuzzolese \and  A. Russo \at
STLab (ISTC-CNR), Via San Martino della Battaglia, 44, 00185 Rome, Italy \\
\email{andreagiovanni.nuzzolese@cnr.it}\\ \email{alessandro.russo@istc.cnr.it}
\and P. Ciancarini \at
DISI, University of Bologna, Mura Anteo Zamboni, 7, 40126 Bologna, Italy \\
Innopolis University, Russian Federation\\
\email{paolo.ciancarini@unibo.it}
\and V. Presutti \at 
LILEC, University of Bologna, Via Cartoleria 5, Bologna, Bologna, Italy \\
STLab (ISTC-CNR), Via San Martino della Battaglia, 44, 00185 Rome, Italy \\
\email{valentina.presutti@unibo.it} 
}

\date{Received: date / Accepted: date}

\maketitle

\begin{abstract}
Social Robotics poses tough challenges to software designers who are required to take care of difficult architectural drivers like acceptability, trust of robots as well as to guarantee that robots establish a personalised interaction with their users.
Moreover, in this context recurrent software design issues such as ensuring interoperability, improving reusability and customizability of software components also arise.

Designing and implementing social robotic software architectures is a time-intensive activity requiring multi-disciplinary expertise: this makes difficult to rapidly develop, customise, and personalise robotic solutions.

These challenges may be mitigated at design time by choosing certain architectural styles, implementing specific architectural patterns and using particular technologies.

Leveraging on our experience in the MARIO project, in this paper we propose a series of principles that social robots may benefit from.
These principles lay also the foundations for the design of a reference software architecture for Social Robots.
The ultimate goal of this work is  to establish a common ground based on a reference software architecture to allow to easily reuse robotic software components in order to rapidly develop, implement, and personalise Social Robots.

\keywords{Social Robots Design  \and Software Architectures \and Architectural Patterns \and Interoperability \and Ontologies \and Linked Open Data}

\end{abstract}

\section{Software Design in Social Robotics}
\label{sec:intro}
\textit{Social Robots}~\cite{Bartneck2004,Breazeal2003,Dautenhahn1998,Duffy1999,Duffy2000, Fong2003} are autonomous embodied agents that interact, collaborate, communicate with humans,  by following the behavioral norms expected by people with whom robots are intended to interact. 
Several definitions have been proposed for the term ``social robot'', but all of them broadly agree that a social robot has the following characteristics:
\begin{enumerate*}[label=\textit{(\roman*)}]
	
	\item \textit{Physical embodiment}, i.e. a social robot has a physical body;
	
	\item \textit{Sociality}, i.e. a social robot is able to interact with people by showing human-like features while following the social rules (defined through society) attached to its role;
	
	\item \textit{Autonomy}, i.e. a social robot makes decisions by itself (the autonomy is sometimes limited in testing phase, like in the Wizard of Oz experimental setting~\cite{Kelley1984,Riek2012}).
	
\end{enumerate*}
Social robots provide valuable solutions for domains, such as  education~\cite{Kennedy2018} or healthcare~\cite{Broekens2009}, where robots must have social skills  to establish and preserve social relationships (even if their domain is dominated by non-social activities)~\cite{Dautenhahn1995}.

\paragraph{Social Drivers.} 
In order to make robots able to establish social relationships, they must be designed so to favor \textit{acceptability}, \textit{trust} and to guarantee a \textit{personalized interaction} with their users.
\textit{Acceptability} is described as the ``the demonstrable willingness within a user group to employ technology for the tasks it is designed to support"~\cite{Dillon2001}.
\textit{Trust} is defined as ``a belief, held by the trustor, that the trustee will act in a manner that mitigates the trustor’s risk in a situation in which the trustor has put its outcomes at risk"~\cite{Wagner2009}.
Humans assess the reliability of a relation when interacting with another agent, hence it becomes critical for robots to act in a way to create and maintain a trusted relation.
No matter how capable is an embodied agent, since its actions may entail risk for its users, they do not interact with it if they do not trust it.
To be accepted to our society robots must show that they are worthy of trust~\cite{Kuipers2018}.

A few studies~\cite{Lee2012a, Gordon2016} have demonstrated that the ability of robots of personalising the interaction with their users is one of the key features that reinforces people's rapport, cooperation, and engagement with a robot.
Robots able to personalise the social interactions adapt their behavior and interaction policy in order to accommodate user preferences, needs, and emotional and psychological state.

Moreover, there exists a variety of challenges that arises in the implementation of a (social) robotics solution~\cite{Ahmad2016}: 
\begin{enumerate*} [label=(\roman*)]
 
    \item How to guarantee the syntactic and semantic interoperability of data exchanged by software artifacts running on a robotic platform?
    
    \item  How to integrate and ease the deployment of  software modules in robotic architecture?
    
    \item How to ease the customization of robot's behavior?
    
    \item How to enhance the reusability of  software components in robot's architectures?
    
\end{enumerate*}
 
\paragraph{Contributions.}
Physical body, hardware and software components contribute to the development of robot social skills and have a role in mitigating these social robotics' challenges,  and establishing and favoring the human-robot interaction~\cite{Dautenhahn2002, Goodrich2008}.
In this paper we will focus on robot's software layer by investigating how to organise software components in order to facilitate the development of robot's social skills as well as to enable robots to carry out a personalised interaction with their users and to increase acceptability, trust, and sociability.
Specifically, we  discuss how the design of the software architecture of the social robots may benefit of well-known architectural patterns and established technologies.
We claim that the design of the robot's software architecture may mitigate  common challenges for social robotics at design time.
Moreover, we propose a reference software architecture for Social Robots that, in our opinion, may favor of the reuse of established technologies and standardised software components thus streamlining the development of social robotics solutions.
This architecture:
\begin{enumerate*}[label=\textit{(\roman*)}]
    
    \item defines a common vocabulary that may facilitate the communications among software designers;
    
    \item serves as a template for developing systems;
    
    \item provides a generalization of 24 existing robotics solutions.
    
\end{enumerate*}
Finally, the architecture is aimed at:
\begin{enumerate*}[label=\textit{(\roman*)}]
    
    \item easing \textit{customizability} and \textit{extensibility} of robot's behavior and social skills; 
    
    \item guaranteeing \textit{explainability} and \textit{predicatability} of robot's decisions;
    
    \item improving both inner (among architectural components) and outer (with external entities) \textit{interoperability};
    
    \item enabling a \textit{rapid prototyping} of robotic applications;
    
    \item enhancing \textit{reusability} of architectural components.
    
\end{enumerate*}

\paragraph{Design Methodology.}
The proposed architecture has been developed with a bottom-up approach.
We have elicited a set of (general) architectural drivers (cf. Section~\ref{sec:requirements}) from the (local) use cases~\cite{MARIOD1.1} defined in the context of  a Socially Assistive Robotics project (cf. Section~\ref{sec:mario})~\cite{Mannion2019}.
This generalization was aimed at formulating the architectural drivers in a way to capture major challenges for  software architecture design in Social Robotics.
The drivers led us to the selection of a set of architectural principles, namely architectural styles and patterns, that Social Robotics architectures may benefit from.
Furthermore,  in order to generalize the design of the architecture we have analysed 24 Social Robotics solutions in order to characterise a reference architecture for a Social Robot.
Finally, drivers and principles have guided the design of the architecture presented in Section~\ref{sec:design}.
As a result, the architecture proposed in Section~\ref{sec:design} is an instance of the reference architecture of a Social Robot discussed in Section~\ref{sec:related} that also follows the architectural drivers elicited in Section~\ref{sec:requirements} and the architectural principles of Section~\ref{sec:choices}.
Finally, this paper discusses the implementation of the proposed architecture in the context of the MARIO project (cf. Section~\ref{sec:mario}).

\paragraph{Outline.}
The rest of the paper is organised as follows. 
An overview of the existing software architectures of the existing social robots is provided in  Section~\ref{sec:related}.
Section~\ref{sec:requirements} discusses the drivers that guided the design of the  architecture.
A set of architectural and technological choices meeting the aforementioned drivers are presented in Section~\ref{sec:choices}.
Section~\ref{sec:design} describes the proposed architecture and Section~\ref{sec:mario} gives an insight of how the architecture has been implemented in MARIO.
Finally, Section~\ref{sec:conclusions} concludes the paper and outlines the future work.
 


\section{Commonalities of Software Architectures for Social Robots}
\label{sec:related}
\begin{figure}[t]
    \centering
    \includegraphics[width=0.6\textwidth]{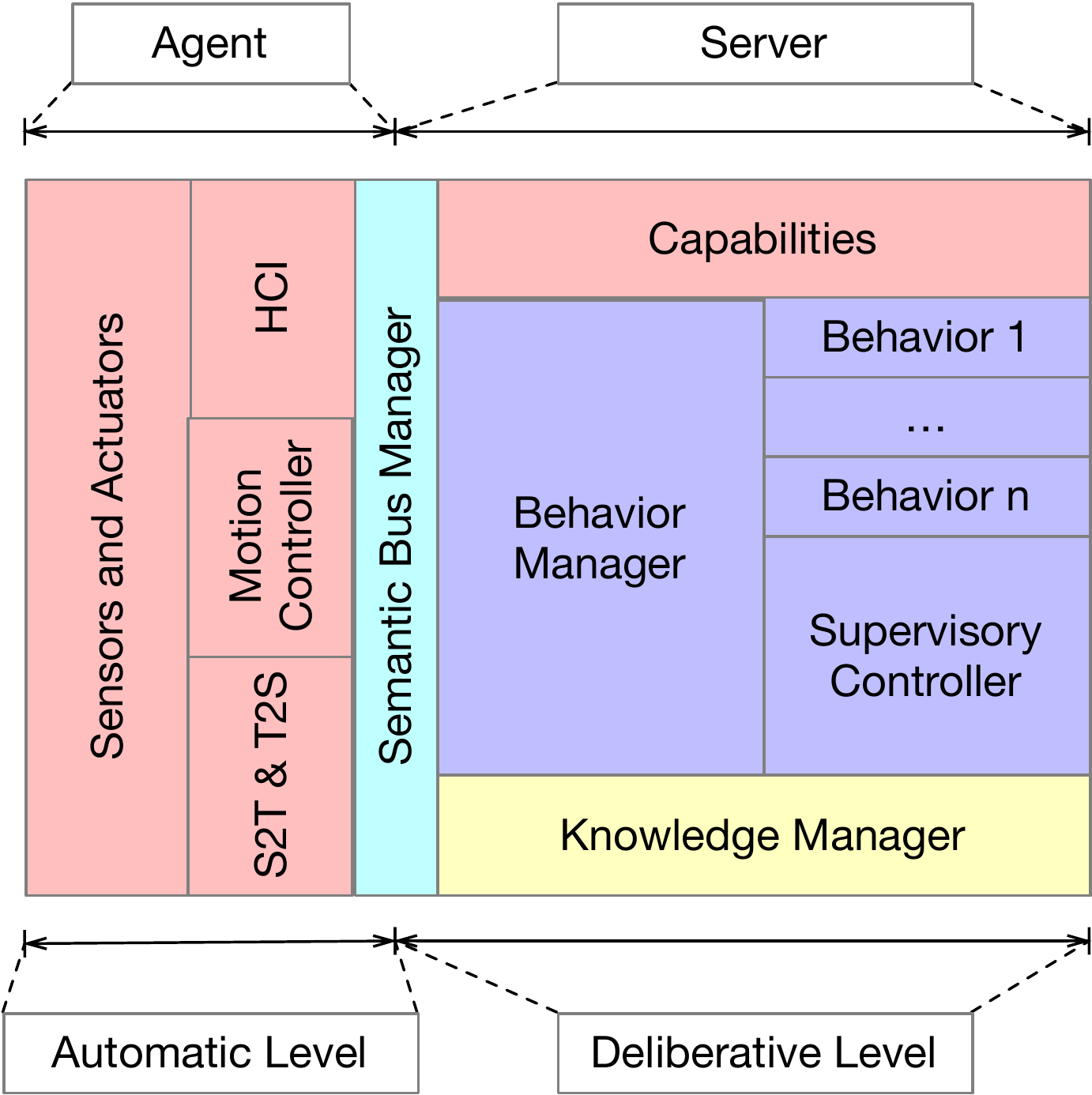}
    \caption{Reference Architecture of a Social Robot.}
    \label{fig:rasr}
\end{figure}

Despite the different application domains and the intended functions, software architectures of Social Robots have a common underlying structure.
This common underlying structure has been synthesized in the Reference Architecture showed in Figure~\ref{fig:rasr}.
This architecture is the result of the analysis of 24 papers describing social robotics systems~\cite{Bonaccorsi2016,Cao2016,Cao2019,Casas2018,Cocsar2020,Fan2017,Dehkordi2015,Fasola2013,Gonzalez-Pacheco2011,Gross2011a,Gross2012,Hirth2011,Jayawardena2016,Kim2009,Louie2014,Louie2020,Mead2010,Portugal2019, Sarabia2011,Shi2018,Torta2014,Uluer2015,Wood2019,Zibafar2019} and it also leverages on our experience in the  MARIO project~\cite{Mannion2019}.



These works have been selected with Systematic Mapping Study (SMS) method~\cite{Petersen2008} that comprises three steps: 
\begin{enumerate}[label=\textit{(\roman*)}]
\item \textit{Planning a study}. The research questions that guided this study have been already discussed and discussed in Section~\ref{sec:intro};

\item \textit{Data collection}. We have enriched the collection of 10 papers identified in a previous analysis~\cite{Asprino2019g}  with a systematic analysis of all the issues of the International Journal of Social Robotics (IJSR).
We analysed IJSR papers in order to select all the papers presenting  social robotics solutions.
This first selection which resulted in 23 papers considered papers' title and abstract only.
A further deeper analysis of these papers was needed in order to keep only papers presenting architectural design of the robot.
This step restricted the set of papers of IJSR papers to 11.
Additional 3 papers were identified by analysing the references of the selected papers.

\item \textit{Analysing the results.} The analysis of the selected 24 papers is presented in this section.

\end{enumerate}

All of the existing architectures define a layering of their components.
Almost all of them differentiate the robot's ``deliberative''   from the ``automatic'' level (terminology borrowed from~\cite{Barber2001}).
The former is meant to decide the next actions the robot has  to perform.
The latter allows the robot to perceive the environment through its sensors (such as lasers, cameras, touch sensors, microphones etc.), to manage the actuators (e.g. wheels engines, speakers etc.) and to provide other basic facilities such as: speech to text (S2T), text to speech (T2S), motion controller and  HCI (the human control interface, i.e. a software component that provides a set of APIs that are meant to manage embodied devices, such as a tablet or buttons, that can be used by users to command the robot).
Software components running on the automatic level are often provided by the standard development kit of  robots.
 
An alternative terminology, proposed by Wood et al.~\cite{Wood2019},  separates sensors, which constitute the ``sense'' layer, from actuators, which form the ``act'' layer, and the deliberative layer is called ``think'' layer.

Although most of the systems  run entirely on robot, there also exist examples of distributed robotic architectures, such as~\cite{Bonaccorsi2016,Kim2009,Louie2014}, in which the deliberative layer is hosted on a remote server and the automatic runs on the robot.
In such architectures the deliberative layer can also control multiple robots at the same time.

Deliberative and automatic layers usually communicate through a semantic bus.
A semantic bus manager is a software component implementing the publish/subscribe pattern for supporting a loosely coupled communication throughout the system (e.g. topics in ROS~\cite{Quigley2009}).
The semantic bus manager allows software component to establish semantic buses that are N-to-N communication channels in which strictly-typed messages flow from the publishers to the subscribers of the channel.

Peculiarities of individual systems apart, all the architectures broadly converge on a  deliberative layer  constituted by  a \textit{Behavior Manager}, a \textit{Knowledge Base Manager}, a set of predefined \textit{Behaviors}, a set of  \textit{Capabilities} (also called \textit{Skills}) and, if necessary, a \textit{Supervisory Controller}.
The Behavior Manager is a software component that gathers information from perceptual components and knowledge base in order to decide the next actions the robot has to perform.
Specifically, it detects from the acquired information the current state of the world and then it activates one of its predefined Behaviors. 
A Behavior is a procedure that makes the robot perform actions (e.g. movements, reproducing sounds etc.).
Examples of Behaviors are entertain the user, reminding something, charging its battery etc.

Some existing architectures~\cite{Cao2019,Gross2011a,Portugal2019} distinguish general purpose functional capabilities of robots from their behavioral capabilities.
In such architectures functional capabilities used by multiple behaviors (such as face detection, person tracking, dialogue managing etc.) form a layer apart that enables the reuse of capabilities among Behaviors. 
These capabilities complement basic functionalities provided (such as S2T, T2S, HCI, Motion controller etc.) usually provided by the agent.

A Knowledge Manager is a software component that provides APIs to store information for supporting the robot's behaviors, tracing the users' activities or preferences, and collecting from the operating environment (e.g. maps).
Many existing architectures~\cite{Cao2016,Cao2019,Cocsar2020,Fan2017,Fasola2013,Kim2009,Louie2014,Uluer2015,Portugal2019} have a centralized Knowledge Base (sometimes articulated into two databases, such as~\cite{Louie2014}) all the components can contribute to/benefit from.

Some architectures, such as~\cite{Cao2016,Cao2019,Fan2017}, explicitly define  a Supervisory Controller that  enables to remotely govern the robot.
Such a interface is part of one of the most common HRI experimental techniques called Wizard of Oz~\cite{Kelley1984,Riek2012}.
In this setting subjects interacting with the robot believe that it is autonomous but it is actually being operated by an unseen human being.

\section{Architectural Drivers}
\label{sec:requirements}
In this section we introduce the main drivers that lead the design of the proposed architecture.
We classified the drivers in functional, presented in Section~\ref{sec:functionalRequirements}, and non-functional, discussed in Section~\ref{sec:nonFunctionalRequirements}. 

\subsection{Functional Drivers}
\label{sec:functionalRequirements}
Functional drivers are defined as capabilities that must be met by a system in order to satisfy a form of request.
Therefore, functional drivers may vary a lot depending on the objectives of the specific system.
In this section we discuss the general functional drivers that a social robot should meet.

\paragraph{Perceiving/Interacting/Motioning within the Environment.} 
A social robot should be able to perceive, to move itself within and physically interact with its operating environment~\cite{Yan2014}.
These drivers \textit{must} be met by all embodied agents that need to interact with their operating environment through their physical body, such as mobile robots or service robots.
However, a social robot may overlook these drivers if the interaction with humans is limited to non-physical languages (e.g. spoken language) and if it does not need to perceive the external environment.
Examples of this kind of robots are the personal assistants (e.g. Amazon Echo, Google Home etc.).

\paragraph{Interacting with Humans.}
Interaction between robots and humans  may  take several forms depending on human-robot proximity (cf.~\cite{Goodrich2008}).
For a social interaction it is important that humans and the robots are co-located in the same environment. 
Within the same environment the interaction may require mobility, physical manipulation, cognitive (e.g. natural language understanding) or emotional (e.g. emotion recognition) abilities.
Here, we stress on the most important abilities that enable interaction with humans and increase the social acceptability of social robots.

\textit{Dialoguing} is a form of interaction where two or more parties communicate.
There are two main forms of human-robot dialogue \textit{verbal} and  \textit{non-verbal}.
social robots should be able to interact with humans using natural language (i.e. verbal communication).
Natural language dialoguing involve capabilities related to speech and natural language processing such as: 
\begin{enumerate*}[label=\textit{(\roman*)}]
	
	\item \textit{Speech recognition}, i.e. the ability of recognizing and translating spoken language into digital-encoded text;
	
	\item \textit{Natural language understanding} (also called machine reading), i.e. the ability of understanding the meaning of the text  and transforming the meaning to a formal structured representation that can be interpreted by machines;
	
	\item \textit{Dialogue managing}, i.e. the ability of  keeping the history and state of a dialog, managing the general flow of the conversation and formulating the semantic representation of the robot's utterances;
	
	\item \textit{Natural language generation}, i.e. the ability of generating natural language  text from a semantic representation of the utterance;
	
	\item \textit{Speech Synthesis}, i.e. the ability of converting the natural language text into speech.
	
\end{enumerate*}

Non-verbal interaction include the use natural cues (e.g. gaze, gestures, body positioning etc.).
The use of basic cues can bootstrap a person's ability of developing a social relationship with a robot~\cite{Duffy2003}.
For example, facial gestures~\cite{Breazeal2000} and motion behaviors~\cite{Duffy2002}  may facilitate to develop a social relationship with a robot.

Emotions play a significant role in  facilitating human-robot interaction (e.g.~\cite{Ogata2000,Canamero2001}).
Therefore, it is important that a social robot is able to recognize and identify emotions in humans, and to express emotions.

\paragraph{Learning and Memorising Knowledge.}
In order to increase its social acceptability and to evolve its social skills, a social robot must be able to learn (e.g. facts, rules, norms etc.)~\cite{Dautenhahn1995b}.
Continuously evolving the robot's knowledge is useful for adapting the robot's behavior in order to accommodate humans' requests in a way they expect.

\subsection{Non-Functional Drivers}
\label{sec:nonFunctionalRequirements}

Non-Functional Drivers specify general properties of a system, rather than its specific functional behavior. 
This section summarizes the general non-functional requirements that a social robot should meet.

\paragraph{Flexibility/Modifiability/Estensibility.}
One of the major challenges in robotics concerns the design of software architectures to support the development of the robot behaviors as plug and play applications~\cite{Chibani2013}.
The robot software architecture should offer an extensibility mechanism to support the composition of new robot behaviors by combinating the existing ones as building blocks.
The requirements of flexibility, modifiability and extensibility of the software architecture is even stronger for social robotics applications.
In fact, social robotics applications might involve a wide variety of components ranging from the component that controls the wheel engines (i.e. components that directly access to the robot's hardware) to the component aimed at understanding what the user says (i.e. components that perform high-level tasks).
Moreover,  extensibility of robot's behavior makes sure that the robot is able to easily extend its capabilities to meet users' requirements, thus increasing its acceptability.

\paragraph{Customizability.} 
Customizing robot's behavior and social skills on the basis of users' needs is crucial in order to improve robot acceptability~\cite{Bailey2009,Fischinger2016}.
Therefore, robotic software applications should either provide an easy customization interface available for behavior's designers or guarantee that the robot is able to autonomously learn how to modify its behavior to meet the user needs.

\paragraph{Predictability. }
To instill trust to its users, a robot must be able to explain and justify its actions, and its behavior should be predictable.
Without a satisfactory explanation for robots' actions, users are not able to assess robots' decisions thus reducing its trustworthiness~\cite{Alaieri2016}.

\paragraph{Interoperability.}
Generally speaking, interoperability is the ability of a system to interact and  work together with other systems~\cite{Wegner1996}.
In order to interoperate, the systems have to agree on a common data format, an unambiguous meaning for the exchanged data, and a protocol for exchanging information.
In robotics we can distinguish two kinds of interoperability: \textit{inner}   and \textit{outer}.
The \textit{inner interoperability} is defined as the ability of a software component to interoperate with another component running on the same robot.
The \textit{outer interoperability} is the ability of a robot  (i.e. one of its software component) to intoperate with another robot~\cite{Mayoral2017}.
Interoperability is not directly related to social robotics challenges but it is a transversal requirement that enables software components to be easily integrated, reused and deployed.

\paragraph{Rapid Prototyping and Reusability of Software Components.}
One of the major problems for the design of a social robot is the definition of the user requirements, which, in turn, is critical to the adoption  of the robot itself.
An accurate early elicitation of the requirements and early feedback on the developing robotic solution may save from costly changes in the robot design at a later stage.
Similar problems are experienced by software developers which usually mitigate these issues with rapid prototyping.
Rapid prototyping refers to a class of techniques that are meant to create and test system prototypes at a very early stage with the aim of getting feedback from users as early as possible.
Rapid prototyping techniques have long been adopted in robotics too (e.g.~\cite{Bartneck2004b,Won2000}) and, since user experience is even more critical than other domain, we believe that rapid prototyping  techniques should be extensively applied in social robotics.
A common strategy to enable rapid prototyping is to compose systems by reusing off-the-shelf software components available on a trusted repository (e.g. apt or yum for Linux).
Therefore, we advocate for the establishment of a repository of software components that can be easily used to quickly build robotic solutions thus letting robot designers to focus only on personalising robot's behavior.
Finally, rapid prototyping is not only a matter of software but also of background knowledge.  
In fact, social robots need a wide range of heterogeneous background knowledge for their tasks which should be available for robots in the same way as software components are available.


	
	
	


\section{Architectural and Technological Choices}
\label{sec:choices}
This section provides an overview of the principles that led the design of the architecture.
Firstly, we chose a robotic approach, namely \textit{behavioral robotics}, which we consider the most appropriate to fulfill the requirements elicited in Section~\ref{sec:requirements}.
Then, we select the architectural style, i.e. service oriented architecture, which best fits with this robotic approach.
Finally, we develop the architecture by applying a set of architectural patterns (i.e. reflection, hot deployment and black board) and by selecting the most suitable technologies with respect to the requirements (semantic web, linked open data).
The remaining of this section briefly presents these principles and motivates such choices.

\paragraph{Behavioral robotics.} \textit{Behavioral robotics}~\cite{Arkin1998} is a robotic approach in which robot's behavior is built in a bottom-up fashion, starting from simple behaviors which are the basic building blocks to realise robot's behavior.
Robot behaviors can run in general simultaneously and are situated (i.e. do not need of complex abstract world models to operate).
Behavior-based robotics is mostly oriented to reactive behaviors in which there is a direct coupling between sensors and actuator.

The architecture presented in this article generalises the classic behavioral robotics approach by combining output of sensors with (symbolic) knowledge of the robot.
Sensors' data and symbolic knowledge are combined in symbolic rules that activate robot's behaviors (not directly actuators).
This strategy increases the customizability and personizability  since the robot's  knowledge base can store users' preferences or habits that can be used to personalise robot's behavior.
Moreover, since robots behaviors are activated by symbolic rules they provide an explanation of the behavior thus favoring the predictability. 
Finally, this approach enhance since  rapid prototyping and reusability since  behaviors, implemented as modular software components, are used as building blocks.

\paragraph{Service-Oriented Computing.} \textit{Service-Oriented Computing (SOC)}~\cite{Papazoglou2003} is a computing paradigm in which services are the fundamentals elements for developing applications.
Services are applications that perform certain functions.
A service is
\begin{enumerate*}[label=\textit{(\roman*)}]
    \item invocable through a platform (e.g. the Web);
    \item self-contained;
    \item technology neutral;
    \item and, its behavior is described by a formal specification.
\end{enumerate*}
Services can be easily composed in order to realise composite functions.
These features make software-oriented architectures an ideal candidate to implement an architecture inspired to behavioral robotics principles.
In this solution robots provide a platform able to host pluggable applications that either realise a robotic behavior or offer some functionality to other applications.
Different benefits might be achieved through the adoption of software oriented principles for designing  robotic software architectures.
\begin{enumerate*}[label=\textit{(\roman*)}]

    \item SOC's neutrality with respect to a specific technology guarantees interoperability of software components.
    
    \item Compositionality of services ensures the rapid development of robot's behaviors.
    
\end{enumerate*}

\paragraph{Hot Deployment.}

Hot-deployable services~\cite{Friese2004} are services that can be added to or removed on-the-fly from a running server.
This mechanism allows to change what is currently deployed on the platform without redeploying the whole platform.
In particular, hot deployment may enable software architectures of social robots to be dinamically extended with new components or to easily customize the deployed ones.

\paragraph{Reflection Pattern.} 
The \textit{Reflection pattern}~\cite{Smith1984} provides a mechanism for dynamically changing structure and behavior of systems. 
This pattern splits a system  into two parts.  
A \textit{meta level} which provides information about system properties that makes the system self-aware of its own structure and behavior.
This information is provided by the so-called \textit{metaobjects}.
A \textit{base level} which implements upon the meta level the application logic by using metaobjects to remain independent with respect to it is likely to change. 
Therefore, any changing in the metaobjects (i.e. changing of the meta level) will be automatically and transparently reflected on the base level.

Behavior controllers of social robots may benefit of a reflection mechanism to dinamically extend capabilities at runtime with no need of re-deploying the whole architecture.
In order to adopt this pattern the behavior controllers need to provide: 
\begin{enumerate*}[label=\textit{(\roman*)}]
    
    \item a \textit{meta level} that should be aware of the robot's capabilities and should allow to dinamically add new robot's abilities (i.e. metaobjects) in form of software modules;

    \item a \textit{base level} that provides an interface for users' requests.
    
\end{enumerate*}
In this scenario user requests' would be transparently handled by using metaobjects. 
This solution would increase the extensibility of the software architecture.

\paragraph{Blackboard Systems.} \textit{Blackboard systems}~\cite{Corkill1991} are systems constituted by independent components that cooperate to solve a problem using a blackboard (i.e. a shared knowledge base) as workplace for developing the solution.
In these systems each component is specialized at solving a certain task of the overall problem (this realizes what is called modularization of expertise).
Components are activated either when a change in the blackboard occurs (e.g. addition or removal of information) or when they receive an external events. 
Components are at same time contributors and beneficiary of the blackboard, namely, their behavior is influenced by the status of blackboard and they record information on the blackboard.

Software architectures of social robots may benefit of a blackboard-system-like design for  several reasons.
With such an architecture software components benefit from and contribute to robot's knowledge hence making robot's behavior conditioned by its knowledge.
Since changing robot's knowledge would affect  its behavior, this kind of architecture would increase the robot's customizability, personizability, predicability and adaptability over the time with respect to user's habits and preferences.
This solution also pushes software components to adopt a shared data model, thus favoring the inner syntactic and semantic interoperability at data level.
Finally, modularization is also enhanced so the rapid prototyping and the reusability of software components.

\paragraph{Semantic Web and Linked Open Data.} 
The \textit{Semantic Web}~\cite{Berners-Lee2001}  is an extension of the Web aims at providing a set of standards that allows data to be shared and reused across application boundaries.
The Semantic Web standards mainly include: \textit{XML} which is a uniform data-exchange format thus providing a common syntax for exchange data; \textit{RDF}\footnote{RDF, W3C Recommendation \url{https://www.w3.org/TR/rdf11-concepts/}} is a framework for  modelling information in form of triples (i.e. subject, predicate, object); \textit{RDFs}\footnote{RDFs, W3C Recommendation \url{https://www.w3.org/TR/rdf-schema/}} provides a data-modelling vocabulary for RDF data; \textit{OWL}\footnote{OWL, W3C Recommendation \url{https://www.w3.org/TR/owl-ref/}} adds constructs for describing properties and classes: among others, relations between classes (e.g. disjointness), cardinality (e.g. ``exactly one"), equality, richer typing of properties, characteristics of properties (e.g. symmetry), and enumerated classes; 
SPARQL\footnote{SPARQL, W3C Recommendation \url{https://www.w3.org/TR/rdf-sparql-query/}} is a query language for retrieving and manipulating data store in RDF format.
Linked Data is structured data that is shared across the internet using  Semantic Web standards.
Linked data distributed with an open licence are called Linked Open Data (LOD).

Robots' architectures can profoundly benefit of \textit{Semantic Web} technologies and \textit{Linked Open Data} paradigm.
\begin{enumerate*}[label=\textit{(\roman*)}]
	
	\item Semantic Web standards allow to easily integrate data generated by the software  components thus favoring the inner  interoperabilty of the software architecture.
	
	\item Semantic Web standards favor the the predictability of the robot.
	
	\item Linked Data  provides a mechanism that allows robots to mutually share knowledge, thus increasing outer interoperability.
	
	\item Linked Open Data paradigm lets to easily reuse existing external datasets so to bootstrap knowledge base with relevant information for robots' activities, and consequently enabling a rapid prototyping of robotic applications.

\end{enumerate*}

\section{Architecture Design}
\label{sec:design}
\begin{figure*}[t]
	\centering
	\includegraphics[width=0.5\textwidth]{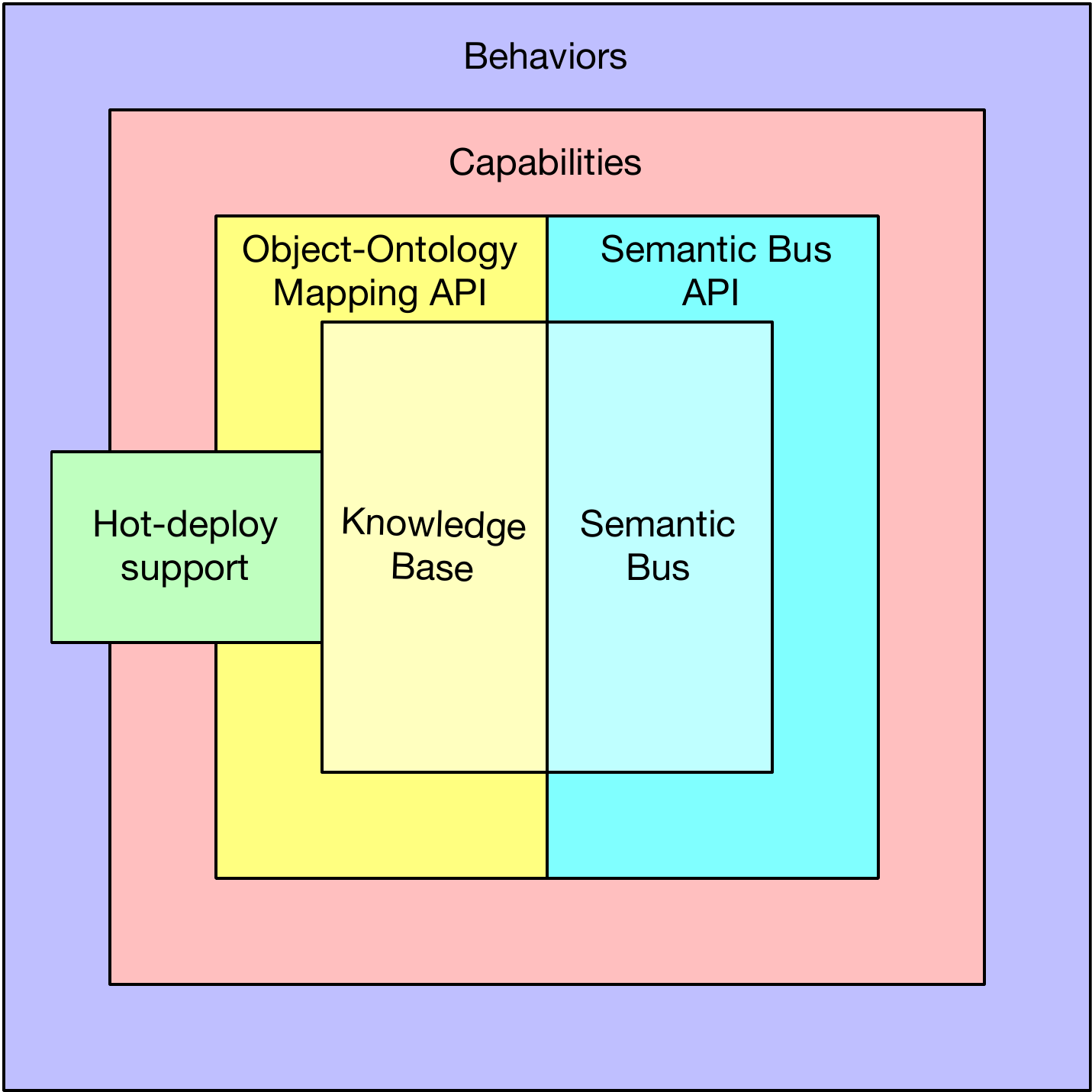}
	\caption{The layered structure of the software architecture. It is worth noticing that the layering organization of the architecture reflects the structure of the architecture presented in Figure~\ref{fig:rasr}. The mapping between the two architectures is emphasized by layers' colors.}
	\label{fig:layers}
\end{figure*}

In order to enhance the separation of concerns among  components, the software architecture  has been organised  into four concentric layers.
These layers are complemented by a cross-layers group of components that provide architecture with hot-deploy support.
Software components on a layer use components of its nested layers.
The outermost layer (i.e. Behaviors) acts as interface with the robot's users and the environment.
Every layer can exist without the layers above it, and requires the layers inside it to function.
Differently, the layer supporting the hot-deploy mechanism is used by its adjacent layers.
Figure~\ref{fig:layers} depicts the architecture layering.
This section provide an overview of the different concerns of the levels.

The base layer is made up of the robot's knowledge  and the bus sub-levels  which constitute the blackboard of the system.
Software components may use knowledge base to store persistent information and the semantic bus to asynchronously exchange messages or to generate and react to events.
Each of these sub-levels is split into two sub-levels, one providing the resources to be accessed (i.e. the knowledge base and the semantic bus) and the other providing the APIs to access to them.
The access to blackboard level is guaranteed by the object-ontology mapping API, which provides software components with facilities for accessing the knowledge base,  and the semantic bus API that allows components to create, subscribe to, or publish messages on a message queue (or topic).

Looking at the higher levels, software components are classified into behaviors and capabilities. 
We define capabilities as software components that give to robots human-like abilities such as: listening, speaking, understanding natural language, moving etc.
These capabilities are typically domain independent and therefore such components can be re-used in different robotic applications. 
Similarly to~\cite{Duffy2000}, capabilities we  classify into basic and convoluted capabilities.
Basic capabilities include both the robot primitive functionalities (e.g. reproducing or capturing sounds) and basic platform services strongly related to the robot primitive functionalities (e.g. speech recognition).
Convoluted capabilities are higher level functionalities (e.g. making phone calls) built on top of the basic ones.
From a developer point of view, both classes of capabilities correspond to functionalities provided by the robot platform.
The main difference between these two classes  is that convoluted capabilities are services that can be included in the robotic platform at run-time using an hot-deploy mechanism, whereas, since typically need of hardware devices, they can only be included at design time. 

Software components that belong to the behavior level are meant to implement the high level  behavior of the robot.
The robot's behaviors are defined as sequences of actions performed by the robot in order to achieve a goal.
An example of behavior is ``entertaining the user''. 
This behavior might involve a series of actions such as: ``approaching'' and ``dialoguing'' to the user, ``showing'' videos, ``reproducing'' music and so on.
Actions requires some robot capabilities like: ``moving'', ``speaking'', ``listening'', ``understanding'', ``showing images'' and so on.

Except for the base layer and the basic capabilities, each level has its own component that allow hot-deploy new components of the same level.
This means that architecture  enables (at running-time)  to deploy new robots behaviors, new robot-capabilities (that do not require new hardware components) and  to extend the structure of the knowledge base by instantiating new ontology-object mappers.
This feature allows to incrementally develop the robot's architecture.

\subsection{Architectural Components}
\label{sec:components}
\begin{figure}
    \centering
    \includegraphics[width=0.95\textwidth]{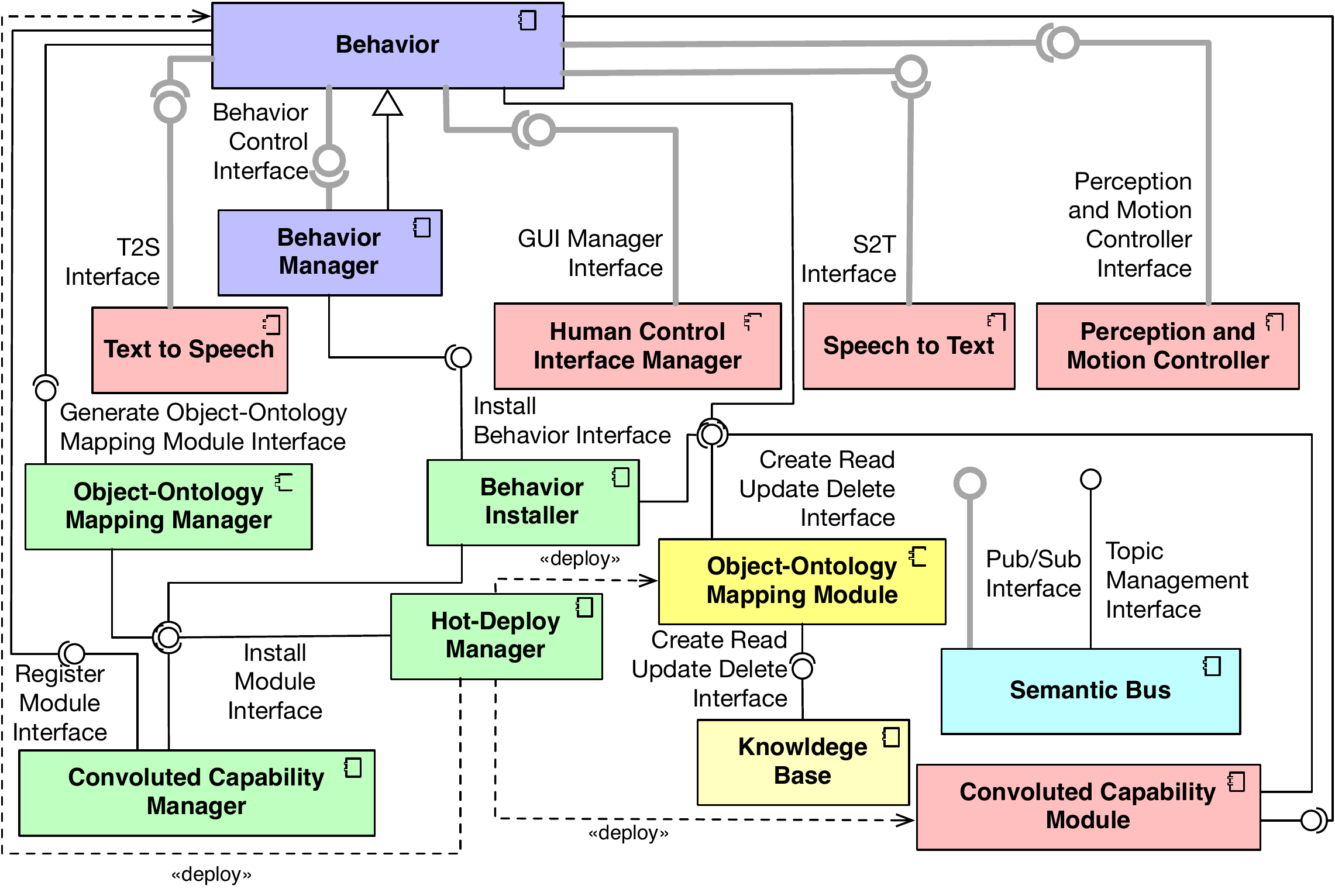}
    \caption{The UML component diagram of the software architecture for social robots. The coloring of the components reflects the color of the layer which the component belongs to, namely: Behavior and Behavior Manager belong to Behaviors level (purple-colored); Text to Speech, Human Control Interface Manager, Speech To Text, and Perception and Motion Controller are part of the Capabilities level (red-colored); Object-Ontology Mapping Module and Knowledge Base are two sub-levels of the knowledge level (yellow-colored) which together with the semantic bus (light-blue-colored) constitute the base level; the software components belonging providing hot-deploy support are green-colored.
    }
    \label{fig:design}
\end{figure}

The Figure~\ref{fig:design} shows the UML component diagram of the proposed architecture.
Each box represents a software component of the architecture.
The assembly connector bridges a component's required interface, which is depicted as a socket, with the provided interface of another component which is represented by a ball.
When the communication between the two components is asynchronous (namely, it is mediated by the Semantic Bus), sockets and ball are gray-highlighted.
The rest of the section is dedicated to the presentation of the design of the architecture and the description  of its software components.

\subsection{Behaviors Layer}

This section gives an overview of the behavior-based mechanism supported by the proposed architecture.
The Behaviors are software artifacts implementing the high-level behavioral capabilities of the robot that are orchestrated and coordinated by the Behavior Manager.
Design and functionalities of these two components is inspired to behavioral robotics principles.

\subsubsection{Robot's Behaviors}
\label{sec:behavior}
A \textit{Behavior} is a software artifact that aims to realize a specific goal. 
Examples of Behavior include  ``entertain the user'', ``locate a user'', ``take user to a place'' etc.
A Behavior relies on perceptual capabilities of the robot that provide sensor data, performs potentially complex processing (e.g. involving retrieving knowledge from  and adding knowledge to the knowledge base), and controls robot’s actuators and devices in order to operate on the environment and interact with the user. 
Each Behavior maintains and updates an internal state, and it decides the actions to perform based on sensor data, its state, the general state of the robot and its internal behavior-specific logic. 

The Behaviors expose an interface (i.e. the \textit{Behavior Control Interface}) that allows the Behavior Manager to control  (i.e. start and stop) them.
Moreover, in order to implement an affordance-based behavior arbitration (the details of this mechanism are provided in the next section), the Behavior Control Interface allows the Behavior Manager to retrieve the situations that can be managed by the Behavior.
Using the Behavior Control Interface, the Task Manager can also grants the access to robot's capabilities to the Behaviors.
Once granted the use to robot's capabilities:
\begin{enumerate*}[label=\textit{(\roman*)}]
	
	\item the Behavior can use the interface provided by the Text To Speech component to make the robot speak;
	
	\item by using HCI Manager Interface provided by the Human Control Interface Manager,  the Behaviors can show to users pictures and videos (if the robot is equipped with a table) or to receive notifications when a button is pressed by a user;
	
	\item the Behavior can subscribe to the Speech to Text topic to retrieve what the user says;
	
	\item the Behavior can use the Perception and Motion Controller Interface to retrieve sensor data, and make the robot move in its operating environment.
	
\end{enumerate*}

The Behaviors are also able to store/retrieve knowledge from the Knowledge Base through an Object-Ontology Mapping Module which is  a RESTful service that mediates CRUD (Create, Read, Update, Delete) operations on the Knowledge Base.
The Object-Ontology Mapping Module makes sure that the interaction with the Knowledge Base complies with the ontology adopted by the Behavior.
A Behavior can also extend (at both intensional and extensional level) the Knowledge Base by generating a new Object-Ontology Mapping Module with the Generate Object-Ontology Mapping Module Interface.
In other words, Behaviors adopt an ontology to structure their data in the Knowledge Base, and, in order to interact with the Knowledge Base they  generate an Object-Ontology Mapping Module that complies with the adopted ontology.
This solution guarantees that Behaviors are decoupled from the technology adopted for the Knowledge Base.

A Behavior can use  robot's convoluted capabilities. 
The architecture depicted in Figure~\ref{fig:design} shows a single Convoluted Capability Module which has to be intended as a placeholder for any possible software components implementing a capability.
This module realizes an interface which is required by the behavior.
This interface provides has to expose capability-specific functionalities that can be used by the Behaviors.
A Behavior can also extend robot's convoluted capabilities by registering new components using the Register Module Interface of the Convoluted Capability Manager.
The new Convoluted Capability Module is then deployed by the Hot-Deploy Manager and becomes available to the other Behaviors.

Since Behaviors communicate with other components either through the Semantic Bus or via REST APIs, there is no restriction on the technology to realise the Behaviors.
This solution increase the modularity of the architecture and, since the architecture allows to deploy applications in any technology, it favors the rapid prototyping of the robot.

Finally, it is worth noticing that  the \textit{Supervisory Controller} (cf. Section~\ref{sec:related}) can be seen as a special Behavior that enable the remote control of the robot.

\subsubsection{Behavior Manager}

The \textit{Behavior Manager} is a special Behavior that actively coordinates other Behaviors and manages  the functional capabilities of the robot. 
Once started Behaviors use robot's computational and sensor/actuator resources granted by the Behavior manager to achieve their goal.
Specifically, the Behavior Manager is responsible for:
\begin{enumerate}
	
	\item Processing incoming data/events and reasoning over the actual state and available knowledge in order to detect situations that require to activate a Behavior;
	
	\item Coordinating, scheduling and prioritizing behavior execution;
	
	\item Activating, suspending, resuming and terminating Behaviors, as a result of a continuous decision making process;
	
	\item Monitoring behaviors executions, to detect successful behavior completions as well as abnormal terminations, failures and exceptions. 
\end{enumerate}
It is worth noticing that Behaviors and their manager implement the reflection pattern where the Behavior Manager is the base level that processes data and events, detects the situation the robot have to act, and activates the Behaviors (i.e. meta-objects).

\paragraph{Behavior Arbitration Mechanism.}
The Behavior Manager implements an hybrid strategy for arbitrating the Behaviors (i.e. deciding which behavior to execute at each time).
It implements a \textit{purely reactive} strategy through a collection of pre-programmed event-condition-action rules.
This strategy targets the most simple requests which do not need to build and reason over a complex, abstract world models.
For example, let the user make a phone call or remembering the user to take his pills does not require a complex control strategy.
The purely reactive strategy has proven to be effective for a variety of problems that can be completely specified at design-time with simple rules~\cite{Mataric1997}.
However, it is inflexible at running time due to its inability to store new information in order to adapt the robot’s behavior on the basis of its experience.
Moreover, the burden of predicting all possible input states and choosing the corresponding output actions is completely left to the designer.

An extension of this purely reactive strategy is a behavior-based approach relying on the notion of \textit{affordance}.
The notion of affordance has been introduced by Gibson~\cite{Gibson1977} who devised a theory of how animals perceive opportunities for actions. 
Gibson called this opportunity \textit{affordance}. 
He suggested that the environment offers the agents (people or animals) opportunities for actions. 
For instance, a door can have the affordance of ``openability''. 
The behavior arbitration strategy implemented by the Behavior Manager exploits and goes beyond the notion of affordance introduced by Gibson.
This mechanism is based on the assumption that not only physical objects, but also complex situations (e.g. the user wants to listen to some music and the robot battery need to be charged) afford  actions.
More precisely, in our model situations afford (i.e. contribute to select)  robot's goals which are then pursued by behaviors which carry out a series of actions to achieve the desired state of the world.

While a situation can be seen as the fullfilment of certain conditions at a certain time, a goal can be interpreted as a certain state of the world the robot desires to achieve.
Both may involve temporal aspects (e.g. lunchtime may afford the task remember the user to take the pills), the perception of certain physical objects, the reception of a command (e.g. I want to listen to some music), or, even the existence of certain state-of-affairs (e.g. the situation the user is sitting on a chair for a long while may afford the task entertain the user).
The Behavior Manager continuously check the current state of the world, and, whenever a condition is satisfied, it retrieves the goals  afforded by the fulfilled situations, it selects one of these goals and it then activates a behavior which synthesizes and execute a plan of how to achieve that goal.
The affordance relation connecting world's situations with robot's goals can be stored in the knowledge base by using the Affordance Ontology Design Pattern~\cite{Asprino2017e}.
This relation can be personalised, i.e. each robot may have its own affordances, and it may be adapted over time to best fit with user preferences.

It is important to note that the arbitration mechanism provides the robot with practical reasoning capabilities~\cite{Bratman1987} (intended as the ``matter of weighing conflicting considerations for and against competing options'').
In fact, a  parallelism with Bratman's BDI (Belief-Desire-Intention) model can be done.
The robot believes the facts stored in the knowledge base and the data received from its sensors.
It uses this information to figure out the current state of the world and to \textit{deliberate} what goals it \textit{desires} to achieve.
Then the robot  commits to the goal to achieve (i.e. intention) and builds a plan to this end.

\paragraph{Hot-Deployment of Behaviors.} Moreover, the Behavior Manager provides the robot's designers with the functionalities for easily hot-deploying Behaviors on the architecture.
Particularly, this component enables the robot's designers to choose which Behaviors (possibly taken from a software repository) to equip the robot with.
The chosen Behaviors are then effectively installed on the architecture by the Hot-Deploy Manager.

It is worth noting that the Behavior Manager guarantees the extensibility, customizability and predictability of the robot's behavior.
The extensibility is guaranteed by the fact that the architecture is open to new Behaviors that can be dynamically added/removed at running time.
In order to customize the robot's behavior a  design just need to modify affordance relations stored in the Knowledge Base.
The robot's behavior can be predicted from affordance relations between situations and Behaviors.

\subsection{Capabilities}

We define capabilities as software components that give to robots human-like abilities such as: listening, speaking, understanding natural language, moving etc.
These capabilities are typically domain independent and therefore such components can be re-used in different robotic applications. 
Similarly to~\cite{Duffy2000}, capabilities we  classify into basic (cf. Section~\ref{sec:basicapabilities}) and convoluted (cf. Section~\ref{sec:convolutedcapabilities}) capabilities.

\subsubsection{Basic Capabilities}
\label{sec:basicapabilities}

Basic capabilities include both the robot primitive functionalities (e.g. reproducing or recording sounds) and basic platform services strictly related to the robot primitive functionalities (e.g. speech recognition).
These capabilities are used by most of the robot Behaviors.
This Section briefly describes the components providing the basic robot capabilities.
The capabilities enabled by these components are the most significant with respect to the  requirements for a social robot (cf. Section~\ref{sec:functionalRequirements} and~\ref{sec:nonFunctionalRequirements}) and that were actually deployed on MARIO (cf. Section~\ref{sec:mario}).
The Section omits components providing general purpose services (e.g. network connectivity).

\paragraph{Text to Speech and Speech to Text.}

The \textit{Text to Speech} (T2S) component aims at converting natural language text into speech.
The T2T implements an interface that allows Behaviors to synthesize and to reproduce synthesized speech.
The \textit{Speech to Text} (S2T) component converts spoken language into digital-encoded text.
The Speech to Text component creates a topic for publishing the converted text and  Behaviors that need to recognize what users say will subscribe to this topic and they will receive a message whenever the text is converted.
Similarly, the Behaviors that need to make the robot speak have to send a message to Text to Speech message queue.
The asynchronous interaction with these tools avoids the Behaviors to busy wait until a new utterance is converted to text (or vice versa) and it decouples the Behaviors from the tools actually used for T2S and S2T.

\paragraph{Human Control Interface Manager.}

Most of the social robots are equipped with some control buttons (e.g. emergency button) and one or more (touch-)screens in order  to complete the message conveyed by verbal communication 
The joint use of verbal and visual language for human computer interaction falls into the broader category of multi-modal human-computer interaction.
This architecture supports a bi-modal interaction involving a both verbal and visual language.
The verbal communication relies on  T2S and  S2T components, whereas  visual communication is ensured by the Graphical User Interface Manager.
The \textit{Human Control Interface (HCI) Manager} component aims at providing Behaviors with facilities for managing the robot's human control interfaces like buttons or tablets.
The component realizes an interface (i.e. the \textit{HCI Manager Interface}) that allows Behaviors to show widgets on the screen or to receive a feedback whenever the user interact with such widgets or a control button.
Similarly to S2T and T2T components, the Behaviors communicate with the HCI Manager through the Semantic Bus.
The use of a standardized HCI Manager within robotics software architecture may standardise the interaction between HCIs and behavior thus increasing the interoperability and  the reusability of software components, and favoring the rapid prototyping.
 
\paragraph{Perception and Motion Controller.}
The \textit{Perception and Motion Controller} provides functional capabilities for supporting human-robot interactions.
It includes a set of software routines that enable the robot to perform a series of motion routines (e.g., approaching the user, following the user, recharging, driving the user to a destination, etc.). 
The Perception and Motion Controller Interface also  provides Behaviors with the access to data coming from several sensors such as:
 \begin{enumerate*}[label=\textit{(\roman*)}]
 	
 	\item \textit{RFID} in order to detect a list of tagged objects;
 	
 	\item \textit{Camera} to detect user using face and posture recognition and extract his relative position and distance;
 	
 	\item \textit{Laser} to perceive and identify dynamic objects/persons that were not included in the static map (SLAM system).
 	
 \end{enumerate*}
Finally, this component provides Behaviors with high-level functionalities such as: go to X (where X is a point within the robot's operating area), give me user's position, give me the tagged objects that are located by the RFID etc.
As for the other components providing basic capabilities, the Perception and Motion Controller  communicates with the Behaviors through the Semantic Bus.
This solution would increase  modularity and  interoperability of the software components thus favoring the rapid prototyping.

\subsection{Convoluted Capabilities}
\label{sec:convolutedcapabilities}

Most of the software applications (not only the robotics ones) that have a  natural language interaction with users rely on (general purpose) NLP services to interpret and to reply to users' utterances.
We call this kind of services Convoluted Capabilities.
These services may require non-negligible computational resources therefore it is desirable to optimise their use as much as possible, in other words, the Behaviors should be enabled to share these services instead of re-instantiating them.
To this end we have replicated at the capability level the same architectural pattern seen at behavioral level involving a component (i.e. the manager) that use the Hot-Deploy manager to dynamically deploy a new software component.
Specifically, Convoluted Capabilities are services that can be dynamically included in the robotic platform at running time.
The new capabilities can be installed by the Behaviors that intend to make available new functionalities for other Behaviors.
The Convoluted Capability Manager is responsible for the dynamic deployment of new capability components.
It realizes an interface (i.e. ``Register Module Interface'') which accepts as input an application module realizing the new capability to deploy.
Once received a module, the Convoluted Capability Manager uses the Hot-Deploy Manager to install the module.

\subsection{Knowledge Management Framework}

This Section presents the architectural components responsible for  the management of the robot's knowledge.
The Knowledge Management Framework consists of two  components, namely the Knowledge Base and the Object-Ontology Mapping Module.

\paragraph{Knowledge Base.}

The \textit{Knowledge Base} is the component intended to store the robot's  knowledge in a structured format.
The reference data model for the Knowledge Base is RDF\footnote{RDF 1.1 Concepts and Abstract Syntax,	\url{https://www.w3.org/TR/rdf11-concepts/}} which is the standard data model adopted in the Semantic Web. 
In our architectural model the Knowlede Base provides facilities that allow to create, read, updated, and delete facts.
The Knowledge Base doesn't have a predefined and fixed schema, but it is defined by the Behaviors.
Finally, the Knowledge Base component also includes a reasoning engine that is able to infer logical consequences (i.e. entailed facts).
It is worth noticing that the adoption of RDF as reference data model increase the interoperability with other Semantic Web compliant systems, and, since it enables the reuse of Linked Open Data datasets, RDF favors also the rapid prototyping.

\paragraph{Object-Ontology Mapping Module.}

An Object-Ontology Mapping Module is a REST service that provides software components with the access to the Knowledge Base.
Given an ontology as input, Object-Ontology Mapping Manager generates an a module that provides  software components with REST APIs for accessing the RDF facts stored in the Knowledge Base in a way that: 
\begin{enumerate*}[label=\textit{(\roman*)}]
    \item reflects the semantics of the ontology, 
    \item and, follows the Object-Oriented paradigm.
\end{enumerate*}
It is easy to recognize the Hot-Deployment pattern already seen for Behaviors and Convoluted Capabilities.
The Object-Ontology Mapping paradigm avoids software components to deal with OWL and RDF or to interact with a knowledge base by means of SPARQL queries.
The Object-Ontology Mapping solution, rather than a direct access to the Knowledge Base, aims to increase the decoupling between the Behaviors and  the Knowledge Base.

\subsection{Semantic Bus}
The \textit{Semantic Bus} is meant to provide the architecture's components with message-based asynchronous communication mechanism that follows the publish-subscribe paradigm.
The Message Broker exposes two interfaces, namely: 
\begin{enumerate*}[label=\textit{(\roman*)}]
	
	\item the \textit{Topic Management Interface} which allows software components to create new topics (also called messages queues);
	
	\item the \textit{Publish/Subscribe Interface} which allows software components to publish messages on/subscribing to a topic.
	
\end{enumerate*}
The asynchronous communication mechanism increases the decoupling among software components.

\subsection{Hot-Deploy Manager}

The \textit{Hot-Deploy Manager} allows to extend the architecture by dynamically deploying new software components.
The Hot-Deploy  Manager aims at providing an OSGi-like\footnote{OSGi, \url{https://www.osgi.org/}} platform for the robot's software architecture enabling a dynamic component model.
In OSGi  applications coming in the form of bundles\footnote{Bundle is a term borrowed from Java-based platforms. A bundle is defined as a group of Java classes and additional resources equipped with a detailed manifest file.} for deployment, can be  installed, started, stopped, updated, and uninstalled without requiring a reboot.
These features ensure the flexibility and the extensibility of the software architecture.

\section{Case Study}
\label{sec:mario}
\begin{figure}[t]
	\centering
	\includegraphics[height=0.4\textheight,keepaspectratio]{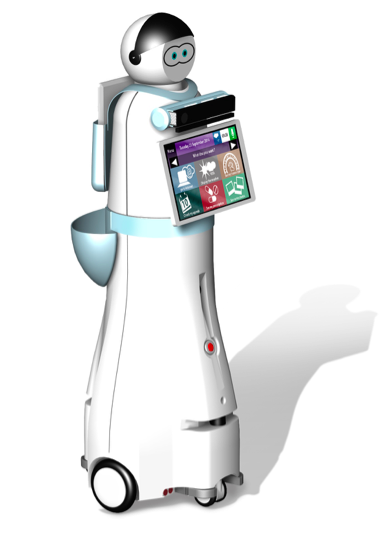}
	\caption{The  Kompa\"{i}-2 robot.}
	\label{fig:kompai2}
\end{figure}

A case study for this work has been provided by the H2020 European Project MARIO\footnote{MARIO project, \url{http://www.mario-project.eu/portal/}}~\cite{Mannion2019}. 
This project has investigated the use of autonomous companion robots as cognitive stimulation tool for people with dementia.
The MARIO robot and its abilities were specifically designed to provide support to people with dementia, their caregivers, and related healthcare professionals.
Among its  abilities, MARIO can help caregivers in the patient assessment process by autonomously performing Comprehensive Geriatric Assessment (CGA) evaluations, it is able to deliver reminiscence therapy through personalized interactive sessions and to entertain its users by playing music or making them reading newspapers or playing videogames.
The overall framework  has been deployed on Kompa\"{i}-2 robots (showed in Figure~\ref{fig:kompai2}), evaluated and validated during supervised trials in different dementia care environments, including a nursing home (Galway, Ireland), community groups (Stockport, UK) and a geriatric unit in hospital settings (San Giovanni Rotondo, Italy).

The architecture proposed in this article is a generalization of the MARIO software architecture.
Specifically, examples of Behaviors in MARIO's architecture are CGA~\cite{Asprino2017b} and Reminiscence~\cite{Asprino2017d} applications.  
These applications use MARIO's Basic Capabilities made available through the integration of commercially available tools such as  Nuance Dragon Naturally Speaking for the Speech to Text Component, Google Speech to Text APIs, and  Karto\footnote{\url{https://wiki.ros.org/karto}} as  motion control system.
Other customised Basic Capabilities were introduced to let Behaviors easily control the robot's GUI interface.
Behaviors share Convoluted Capabilities providing common NLP services (e.g. Stanford's CoreNLP~\cite{Manning2014}) for dialoguing with users~\cite{Russo2019}.
Behaviors and Capabilities rely on a Knowledge Management Framework~\cite{Asprino2019g,Asprino2017c} consisting of a triple store (accessible through Apache Jena\footnote{\url{https://jena.apache.org/}} and initially populated with Linguistic and Common Sense Knowledge~\cite{Asprino2018a,Gangemi2016}) and an Object-Ontology Mapper called Lizard\footnote{\url{https://github.com/anuzzolese/lizard}}~\cite{Asprino2019g}.
Finally, the Semantic Bus is based on the Event Admin Service provided by Apache Felix\footnote{\url{https://felix.apache.org/}} which also provides the OSGi implementation that enables the Hot-Deploy mechanism.

\section{Conclusions and Future Work}
\label{sec:conclusions}
This paper has focussed on how a design based on a reference software architecture can mitigate challenges in social robotics, such as improving acceptability, trust of robots, and guaranteeing that robots establish a personalised relationship with their users.
We have adopted a bottom-up approach for defining general architectural requirements starting from specific use cases defined in the context of MARIO, a Socially Assistive Robotics project funded by the EU, and then generalizing them.
These requirements led us to the selection of a set of technologies, architectural styles and patterns, that Social Robotics architectures may benefit from.
We remark that non-functional requirements and principles have guided the design of the architecture.
The reference architecture introduces  a series of standardised software components that are meant to increase the extensibility, customizability, predicatability, interoperability of the software architecture as well as to favor the rapid prototying of the Social Robotics solutions.

The ultimate goal of this work is to establish a common open-ended software architecture  so to encourage the development of standardised software components whose Social Robots can benefit from.
As shown by the use case presented in Section~\ref{sec:mario}, many of these components can be provided by off-the-shelf software tools or  are available as research prototypes.
However, even if these components are available, they need to be adapted (or wrapped) in order to be integrated in the architecture (to this end the Adapter Pattern~\cite{Gamma1994} can be considered).
Others, especially the components responsible for the hot-deployment mechanism, need to be implemented from scratch.
In addition to the development of new components, it is also necessary to define a set of standard protocols that let components communicate.
On the basis of the experience in the presented case study we are  implementing a proof-of-concept of the architecture.
Particularly, we are generalising protocols used in the MARIO architecture and we are developing an affordance-inspired Behavior Manager.
Finally, we plan to explore the possibility of enhancing the autonomy of robot by design.
To this end an autonomic computing solution~\cite{Kephart2003} is being considered
in order to endow the robot with self-configuration, self-optimization, self-healing and self-protection capabilities.

\begin{acknowledgements}
The research leading to these results has received funding from
the European Union Horizons 2020 the Framework Programme for Research and Innovation (2014-2020) under grant agreement 643808 Project MARIO ``Managing active
and healthy aging with use of caring service robot''.
\end{acknowledgements}

\printbibliography
\end{document}